\documentclass[sigconf]{acmart}

\AtBeginDocument{
  }

\copyrightyear{2026}
\acmYear{2026}
\setcopyright{cc}
\setcctype{by}
\acmConference[KDD '26]{Proceedings of the 32nd ACM SIGKDD Conference on Knowledge Discovery and Data Mining V.2}{August 09--13, 2026}{Jeju Island, Republic of Korea}
\acmBooktitle{Proceedings of the 32nd ACM SIGKDD Conference on Knowledge Discovery and Data Mining V.2 (KDD '26), August 09--13, 2026, Jeju Island, Republic of Korea}
\acmDOI{10.1145/3770855.3818933}
\acmISBN{979-8-4007-2259-2/2026/08}

\usepackage{array}
\usepackage{multirow}
\usepackage{pgfplots}
\usepackage[dvipsnames]{xcolor}
\usepackage{makecell}
\usepackage{pifont}
\newcommand{\cmark}{\ding{51}} 
\newcommand{\xmark}{\ding{55}}

\begin{document}

\title{MammoExpert: Benchmarking Chain-of-Thought Reasoning in Mammography Diagnosis}

\author{Di Dai}
\authornote{Both authors contributed equally to this research.}
\orcid{0009-0002-1832-5164}
\affiliation{
  \institution{
State Key Laboratory of General Artificial Intelligence, School of Intelligence Science and Technology, Peking University}
  \city{Beijing}
  \country{China}
}
\email{didai@stu.pku.edu.cn}

\author{Bo Liu}
\authornotemark[1]
\affiliation{
  \institution{State Key Laboratory of General Artificial Intelligence, School of Intelligence Science and Technology, Peking University}
  \city{Beijing}
  \country{China}
}
\email{liubo2022@stu.pku.edu.cn}

\author{Youcheng Li}
\authornotemark[1]
\affiliation{
  \institution{State Key Laboratory of General Artificial Intelligence, School of Intelligence Science and Technology, Peking University}
  \city{Beijing}
  \country{China}
}
\email{youchengli@stu.pku.edu.cn}

\author{Haojun Yu}
\authornotemark[1]
\affiliation{
  \institution{State Key Laboratory of General Artificial Intelligence, School of Intelligence Science and Technology, Peking University}
  \city{Beijing}
  \country{China}
}
\email{haojunyu@pku.edu.cn}

\author{Zhuohang Bian}
\affiliation{
  \institution{School of Computer Science and Engineer, Beijing University of Aeronautics and Astronautics}
  \city{Beijing}
  \country{China}}
\email{22373017@buaa.edu.cn}

\author{Quanlin Wu}
\affiliation{
  \institution{Center for Data Science, Peking University }
  \city{Beijing}
  \country{China}
}
\email{quanlin@pku.edu.cn}

\author{Dong Wang}
\affiliation{
  \institution{Yizhun co. ltd}
  \city{Beijing}
  \country{China}
}
\email{dong.wang@yizhun-ai.com}

\author{Sichen Meng}
\affiliation{%
 \institution{International School, Beijing University of Post and Telecommunications}
 \city{Beijing}
 \country{China}}
 \email{2022213422@bupt.cn}

\author{Hongye Xuan}
\affiliation{
  \institution{School of Public Health, University of Michigan, Ann Arbor}
  \city{Beijing}
  \country{China}}
\email{xhongye@umich.edu}

\author{Zijie Lan}
\affiliation{
  \institution{Future Technology College, Xi'an Jiaotong University}
  \city{Xi'an}
  \country{China}}
\email{16681914477@stu.xjtu.edu.cn}

\author{Shenda Hong}
\authornote{Corresponding author.}
\affiliation{
  \institution{Peking University}
  \city{Beijing}
  \country{China}}
\email{hongshenda@pku.edu.cn,}

\author{Liwei Wang}
\authornotemark[2]
\affiliation{
  \institution{State Key Laboratory of General Artificial Intelligence, School of Intelligence Science and Technology, Peking University}
  \city{Beijing}
  \country{China}}
\email{wanglw@cis.pku.edu.cn}

\renewcommand{\shortauthors}{Di Dai et al.}

\begin{abstract}
Mammography is an essential tool for breast cancer detection, with millions of examinations conducted annually. However, publicly available high-quality mammography datasets for AI development remain limited in both scale and annotation richness, particularly regarding pathological subtype coverage and structured diagnostic reasoning annotations. In this paper, we present MammoExpert, the first mammography dataset with Chain-of-Thought reasoning annotations across three diagnostic phases: (i) primal observation, (ii) factual assessment, and (iii) diagnostic synthesis. 
Comprising 2,379 mammography images covering 67 WHO-classified histopathology subtypes, each exam provides 42 radiographic features annotated by nine senior radiologists. We evaluate its performance on the breast lesion
classification task, demonstrating superior accuracy
and reasonability compared to existing classification models.
Combining public dataset CBIS-DDSM with MammoExpert yields 7.1\% classification accuracy improvement, while the training model to learn CoT reasoning achieves another 4\% gain on the MammoExpert test set. Similar improvements are observed on INBreast and Vindr datasets, where the full approach yields accuracy gains of 6.9\% and 6.7\%, respectively. MammoExpert can serve as a benchmark for interpretable breast lesion diagnosis through explicit CoT reasoning. The dataset, code, and documentation are available at \url{https://github.com/Ericdd90/MammoExpert.}
\end{abstract}




\maketitle

\section{Introduction}
\label{sec1}
Breast cancer remains a significant threat to women health, causing 670,000 deaths per year~\cite{kim2025global}. 
Mammography is a recommended imaging for accurate diagnosis of breast lesions, which can significantly reduce mortality~\cite{gotzsche2013screening}.
Expert physicians perform two complementary approaches to mammography-based diagnosis: (1) pattern recognition for typical cases; and (2) probabilistic diagnostic reasoning for suspicious cases~\cite{gotzsche2000screening}.
While AI systems have achieved great success in mammography-based diagnosis, they only perform pattern recognition, but cannot perform reasoning processes like human experts. 
This not only limits the interpretability during human-AI interaction, but also bounds the capabilities of AI systems to analyze hard cases.

\begin{figure*}[!t]
    \centering
    \includegraphics[width=\textwidth]{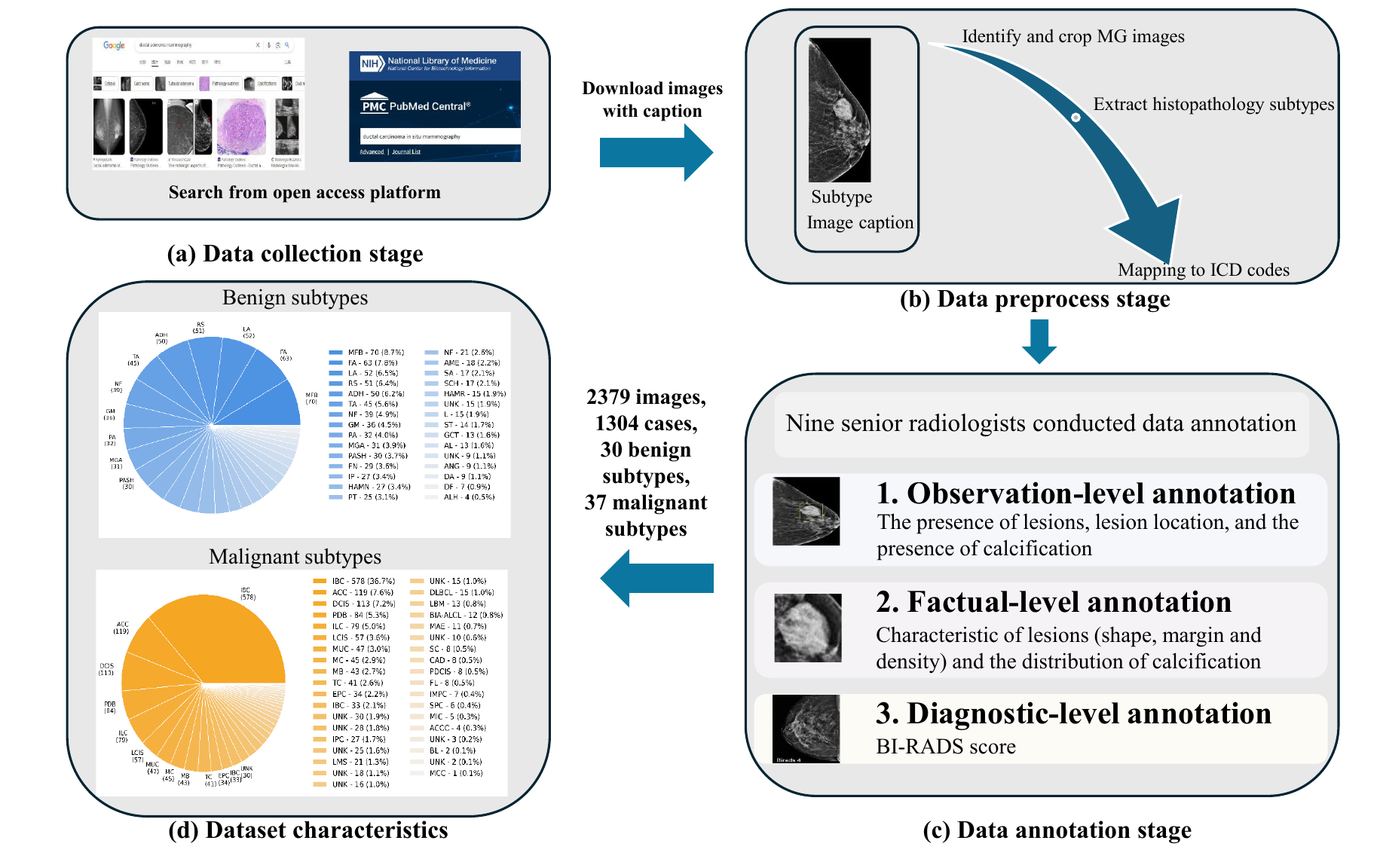}
\caption{
\textbf{Construction of MammoExpert dataset.} 
(a) \textbf{Collection}: 2,379 mammograms from Google Image/PubMed using histopathological criteria. 
(b) \textbf{Preprocessing}: Modality verification and metadata extraction. 
(c) \textbf{Annotation}: Three-phase labeling (observation $\rightarrow$ morphological analysis $\rightarrow$ diagnostic synthesis) by senior radiologists, with multidisciplinary discordance resolution. 
(d) \textbf{Statistics}: Balanced subtype coverage (30 benign vs. 37 malignant) across 67 categories.
}
    \label{fig_data_statistics}
\end{figure*}

With recent advances of large language models (LLM)~\cite{achiam2023gpt,guo2025deepseek}, AI systems can now perform CoT reasoning and solve complicated problems~\cite{wei2022chain,miao2024chain}.
To facilitate reasoning in the medical-AI community, we provide MammoExpert, a mammography diagnosis benchmark capturing the full diagnostic reasoning process of expert physicians, as shown in Figure~\ref{fig_data_statistics}.
The reasoning process consists of three key stages: (i) primal observation (lesion/calcification detection), (ii) factual assessment (lesion/calcification analysis), and (iii) diagnostic synthesis (assigning cancer risk scores)~\cite{sinn2013brief,mcdonald2016clinical}.
The comprehensive MammoExpert benchmark contains 2,379 images, covering 67 histopathology of breast lesions defined by WHO-standard~\cite{who_classification_breast_tumours,sinn2013brief} and 42 measurable features defined by BI-RADS standard~\cite{spak2017bi}. Each case is annotated by nine senior physicians with 15 $\pm$ 6 years' experience.

Additionally, we develop a baseline model and conduct a comprehensive evaluation of its
performance on the classification task. This analysis establishes a fair and consistent framework for evaluating  performance. We anticipate that the MammoExpert dataset and baseline model will drive the widespread adoption of breast lesion classification in clinical practice, accelerating the healthcare industry’s transition toward intelligence and reasonable. Extensive experiments demonstrate that the MammoExpert dataset and the CoT reasoning annotations significantly enhance model capabilities and interpretability.
When training on CBIS-DDSM~\cite{sawyer2016curated} and MammoExpert, our model achieves a 7.1\% absolute improvement over baselines trained on CBIS-DDSM for the pathology classification task.
On the MammoExpert test set, explicit modeling diagnostic CoT reasoning process yields 4\% accuracy gains comparing with directly providing predictions. Similar improvements are observed on INBreast
and Vindr datasets, where the full approach yields accuracy gains
of 6.9\% and 6.7\%, respectively. These improvements come from carefully analyzing multiple intricate features and synthesizing conclusions based on the gathered evidence.

The contribution of this paper is three-fold. 

(1) We build open-source MammoExpert, a diagnostic reasoning benchmark with comprehensive data covering 67 WHO-classified histopathology types. 

(2) We propose a three-stage Chain-of-Thought reasoning process that derived real-world clinical workflows,  and evaluate its performance on the breast lesion
classification task, demonstrating superior accuracy and reasonability compared to existing classification models. 

(3) We conduct proof-of-concept experiments to demonstrate that AI systems can provide better predictions with reasoning processes.
The MammoExpert has the potential to facilitate future research on AI-based precision and trustworthy healthcare.

The rest of this paper is organized as follows. We first review the related work in Section~\ref{sec2}. Then,
Section~\ref{sec3} describes the mammography dataset in detail. After that, we present the proposed Chain-of-Thought model in Section~\ref{sec4} and report experiments in Section~\ref{sec5}, respectivel. Section~\ref{sec6} discusses limitations and ethical considerations. Finally, a brief conclusion of this paper is summarized in Section~\ref{sec7}, and Section~\ref{sec8} provides a disclosure regarding the use of generative AI tools in this work.

\section{Related Work}
\label{sec2}
\subsection{Chain-of-Thought Reasoning}
Chain-of-Thought reasoning has emerged as a powerful paradigm for improving complex problem-solving in large language models (LLMs), enabling them to decompose tasks into intermediate reasoning steps rather than relying solely on shallow pattern matching~\cite{wei2022chain,kojima2022large,yao2023tree}. CoT has proven particularly effective in domains such as mathematical problem solving, commonsense reasoning, and multi-hop question answering, where models benefit from articulating step-by-step logical processes~\cite{zhang2022automatic,wang2023self}.

This approach aligns naturally with medical diagnosis, where clinicians frequently employ sequential reasoning to integrate diverse pieces of evidence—such as imaging features, clinical history, and statistical probabilities—before reaching a conclusion~\cite{guyatt1993users}. For example, radiologists interpret mammograms by systematically evaluating specific features (e.g., mass shape, margins, density) and mapping these observations to potential diagnostic categories through structured mental workflows~\cite{lehman2017national}.

Despite these parallels, many contemporary medical vision systems remain predominantly end-to-end models, mapping images directly to diagnostic predictions without modeling intermediate reasoning steps. This black-box approach may compromise interpretability, hinder clinical trust, and lead to suboptimal performance on complex or ambiguous cases where nuanced feature-based reasoning is crucial~\cite{holzinger2017we}.

MammoExpert seeks to address this gap by introducing structured annotations that explicitly model radiologists’ diagnostic reasoning. Inspired by CoT principles, MammoExpert captures the radiologist’s three-phase interpretive workflow: detecting salient mammographic features, characterizing these features in detail, and correlating them with diagnostic categories. By aligning model architecture and training with human reasoning patterns, this strategy aims to enhance both diagnostic accuracy and model explainability, ultimately fostering safer and more trustworthy AI deployment in breast imaging.

\subsection{Mammography Datasets}

As more and more mammography datasets with diverse characteristics are being constructed, the use of mammography for assisting breast cancer diagnosis has attracted widespread attention. Some of these datasets are publicly available to researchers, while others are restricted-access or non-public, as shown in Table \ref{table111}.

\begin{table}[!t]
\centering
\footnotesize
\caption{Existing Public Mammography Datasets}
\label{table111}
\resizebox{\linewidth}{!}{
\begin{tabular}{ccccc}
\toprule
\textbf{Dataset} & \textbf{Year} & \textbf{Images} & \textbf{Annotation} & \textbf{BI-RADS} \\
\hline
MIAS \cite{suckling1994mammographic} & 1994 & 322 & Circle around the
finding & No \\
DDSM \cite{heath1998current} & 1999 & 10,480 & Contour enclosing the
finding & Yes \\
CBIS-DDSM \cite{lee2017curated} & 2017 & 3,103 & Contour enclosing the
findings & Yes \\
INBreast \cite{moreira2012inbreast} & 2012 & 410 & Contour enclosing the
finding & Yes \\
NYU Dataset \cite{wu2019deep} & 2019 & 1,001,093 & Contour enclosing the
finding & Yes \\
CSAW-CC \cite{dembrower2020multi} & 2020 & 98,788 & Contour enclosing the
finding & No \\
OMI-DB \cite{halling2020optimam} & 2021 & 3,072,878 & Rectangular region of
interest & No \\
CMMD \cite{cui2021chinese} & 2021 & 5,202 & No local annotation & No \\
VinDr-Mammo \cite{nguyen2023vindr} & 2022 & 20,000 & Rectangular region of
interest & Yes \\
Mammo-Bench \cite{bhole2025mammo} & 2025 & 19,731 & Region of interest mask
 & Yes \\
\bottomrule
\end{tabular}
}
\end{table}

The DDSM \cite{heath1998current} and MIAS datasets \cite{suckling1994mammographic} were the earliest publicly available digitized screening mammography datasets that provided precise annotations of breast abnormalities (i.e., screen-film mammography). The MIAS dataset was released in 1994 and contains 161 studies collected in the UK, while the DDSM dataset was collected from multiple institutions in the United States and contains 2,620 examinations. Compared with MIAS, the DDSM dataset is significantly larger in scale and follows the BI-RADS standard. The INbreast dataset \cite{moreira2012inbreast}, released in 2012, was the first publicly available digital mammography dataset providing lesion annotations and overall examination assessments, all compliant with BI-RADS. In 2019, the NYU dataset \cite{wu2019deep} was released, consisting of 229,426 exams from 141,473 screening women at NYU Langone Health, including 1,001,093 images. This dataset includes breast cancer diagnoses based on biopsy results, BI-RADS assessments, breast density information, and biopsy lesion annotations. Although this dataset has not been publicly released, subsequent research based on it has demonstrated that large-scale mammography datasets can support computer-aided systems and improve radiologists’ performance. Around the same time, the CSAW-CC dataset \cite{dembrower2020multi} was released publicly for evaluating AI tools in breast cancer applications. It comprises 1,303 cancer cases and 10,000 randomly selected control samples from Karolinska University Hospital. The CSAW-CC dataset is a subset of the full CSAW dataset, including women screened in the Stockholm region between 2008 and 2015. In the cancer cases, tumors visible on mammograms were manually annotated at the pixel level. Another large-scale dataset is the OPTIMAM database OMI-DB \cite{halling2020optimam}, which includes images and clinical data from 172,282 women screened and diagnosed across multiple institutions in the UK since 2011. To access the OMI-DB dataset, research teams must submit an application specifying their scientific research objectives using the dataset, which will be reviewed by the OPTIMAM Steering Committee. Additionally, a recently released Chinese mammography database includes 1,775 studies from multiple institutions in China. All cases involve biopsy-confirmed benign or malignant breast lesions, among which 749 cases include molecular subtype information. The VinDr-Mammo dataset \cite{nguyen2023vindr} is a publicly available large-scale mammography dataset containing 5,000 four-view examinations, which contains 20,000 images in total. This dataset was collected from two major hospitals in Hanoi, Vietnam, with all image annotations following the BI-RADS reporting standard. Recently, the MammoBench dataset~\cite{bhole2025mammo} was introduced as a large-scale, standardized benchmark for mammography. It contains 19,731 images, which provides comprehensive annotations including BI-RADS assessments, breast density, abnormality types, and ROI masks. 

Existing large-scale datasets like CBIS-DDSM~\cite{lee2017curated} and VinDr-Mammo~\cite{nguyen2023vindr} provide biopsy-confirmed mammograms but lack reasoning annotations, limiting AI systems' capacity for diagnostic reasoning. Also, these resources omit critical text-based medical knowledge, such as lesion features and fine-grained histopathological information~\cite{degrave2023dissection}. MammoExpert addresses these gaps by incorporating 42 radiographic features per case across 67 WHO-classified histopathological subtypes, including rare variants. This structured knowledge base enables models to leverage expert-curated lesion characteristics and diagnostic logic during inference, facilitating clinically-grounded reasoning pathways absent in prior datasets.

\section{Dataset}
\label{sec3}
MammoExpert is a mammography dataset containing diagnostic reasoning annotations across 2,379 images covering 67 WHO-defined histopathological subtypes, which is shown in Figure~\ref{fig_data_statistics}. According to histopathological diagnosis and following the WHO Classification of Tumours (WHO23), the dataset includes 30 benign subtypes and 37 malignant subtypes.
Such fine-grained pathological categorization enables detailed analysis of subtype-level diagnostic performance and supports learning under long-tailed data distributions.

Each case is associated with comprehensive annotations, including lesion characteristics, structured ultrasound reports, BI-RADS assessments, and pathological labels, which is shown in Figure~\ref{fig1}.
The overall dataset composition is summarized in Table~\ref{tab:dataset_summary}, while The detailed subtype distributions are illustrated in Figure~\ref{fig_data_statistics}.
Nine senior radiologists annotate 42 radiographic image features per case, calcification patterns, and BI-RADS scores, enabling AI systems to replicate clinical reasoning from imaging features to pathological diagnosis.

\begin{figure}[!t]
\centering
\footnotesize
\label{fig4}
% 左：JSON 结构
\begin{minipage}[t]{0.5\linewidth}
\textbf{MammoExpert\_dataset.json}
\begin{verbatim}
{
  "00001": {
    dataset
    modality
    source_information
      |-- url
      `-- license
    patient_id
    age
    lesion_mask_file
    device_type
    side
    pathology_histology
      |-- pathology
      `-- subtype
    lesion_box
      `-- lesion_id: [x1, y1, x2, y2]
    caption
      |-- image_relative_caption
      |-- image_full_caption
      `-- key_words
    image_quality
    annotation
      |-- BI_RADS
      |-- Breast_Composition
      |-- Mass_Shape
      |-- Lesion_Margin
      |-- Calcification_Pattern
      `-- ...
  }
}
\end{verbatim}
\end{minipage}
\hfill
% 右：字段说明
\begin{minipage}[t]{0.45\linewidth}
\textbf{Field Description}
\begin{itemize}
  \item \textbf{dataset}: Identifier of the data source.
  \item \textbf{modality}: Imaging modality.
  \item \textbf{source\_information}: Source URL and license.
  \item \textbf{patient\_id}: Anonymized patient identifier.
  \item \textbf{age}: Patient age ($-1$ if unavailable).
  \item \textbf{pathology\_histology}: Pathology and subtype.
  \item \textbf{lesion\_box}: Bounding box $[x_1,y_1,x_2,y_2]$.
  \item \textbf{caption}: Image-level textual descriptions.
  \item \textbf{annotation}: Radiological attributes (e.g., BI-RADS).
\end{itemize}
\end{minipage}
\caption{Structure of the MammoExpert dataset annotation file.}
\label{fig1}
\end{figure}

\begin{table}[!t]
\centering
\footnotesize
\caption{Dataset Summary of MammoExpert by Pathology.}
\label{tab:dataset_summary}
\setlength{\tabcolsep}{10pt}
\begin{tabular}{lccc}
\toprule
\textbf{Pathology} & \textbf{Images} & \textbf{Patients} & \textbf{Subtype Count} \\
\midrule
Benign      & 803 & 364 & 30 \\
\midrule
Malignant   &    1576                     &     940                    & 37 \\
\midrule
Total       & 2,379 & 1,304 & 67 \\
\bottomrule
\end{tabular}
\end{table}

\subsection{Data Collection}

We collect mammographic images and their associated case reports from publicly available sources, including Google Images\footnote{https://images.google.com/}, PubMed\footnote{https://pubmed.ncbi.nlm.nih.gov/}, and other open-access medical resources. The search is guided by histopathological subtype names from the WHO classification of breast tumours~\cite{who_classification_breast_tumours}, which are used as query terms to identify cases with imaging evidence and diagnostic descriptions. For each candidate case, the linked report or source page is reviewed to confirm that the image corresponded to a breast lesion and that the accompanying text provided sufficient pathological information. Figure~\ref{fig_data_statistics}(a) shows the geographic distribution of the collected cases and the platforms from which they are obtained. Inclusion criteria require biopsy-confirmed diagnoses, which are applied to ensure that each image can be linked to a reliable histopathological label rather than a purely radiological impression. In total, the dataset covered 67 histologically validated subtypes. These subtypes are further mapped to their corresponding ICD codes~\cite{o2005measuring} through expert manual annotation, which allows the image-level labels to be aligned with standardized disease categories.

\subsection{Data Preprocessing}

Our automated pipeline initially identifies 4,382 candidate images, of which 2,003 (45.7\%) are systematically discarded through two-phase quality assurance where 412 images during initial screening for non-conforming modalities (ultrasound/MRI cross-sections, diagrammatic illustrations), and 1,591 images in secondary verification due to incomplete pathology annotations or technical artifacts. All images are fully
anonymized, with no patient privacy information, personal identifiers and protected health information
(PHI), which are used solely for non-commercial research purposes. We conduct strict duplicate removal via
image hashing and visual verification. The image formats and intensity scales are unified without artificial
post-processing or modification. All annotations are newly produced by experienced radiologists based on
original visual content. This yields 2,379 diagnostic-quality images meeting inclusion criteria for downstream analysis. Mammographic regions are cropped to avoid interruption from the background. From description texts, we also extract histopathology categories to support medical prior knowledge and mapping them into ICD codes by expert manual annotation, as shown in Figure~\ref{fig_data_statistics}(b). This process preserves the clinically relevant visual information in the mammograms and reduces noise from background regions or technical artifacts.

\subsection{Data Annotation}
Nine senior radiologists with 15~$\pm$~6 years of experience annotate each mammogram in three phases. First, they document basic observations by checking for visible lesions, marking their locations, and noting any calcification presence. Next, they record detailed characteristics including lesion shape, margin type, density level, and calcification distribution patterns. Finally, they provide diagnostic conclusions by assigning BI-RADS scores and matching cases to specific cancer subtypes. Figure~\ref{fig_data_statistics}(c) illustrates this workflow. All annotations undergo quality review by an expert panel, which resolve disagreements through discussion for 6.8\% of cases to ensure consistency while preserving clinically realistic variations.

The annotation process is conducted on a proprietary web-based platform designed for medical image labeling. Each radiologist is assigned 150-200 unique cases to annotate independently, ensuring no overlap between experts. Following initial annotations, a separate two-person review team performs quality checks on the platform. For cases with disputed annotations, the review team directly consults the original annotator through built-in discussion tools to reach a consensus. We calculate the Fleiss' kappa coefficient to quantify the overall agreement among the three radiologists, and the result is 0.82, which indicates excellent inter-annotator consistency according to standard medical annotation guidelines. This quantitative evidence confirms the high reliability of the annotations in our MammoExpert dataset.

\begin{figure}[!t]  
    \centering
    \includegraphics[width=\columnwidth]{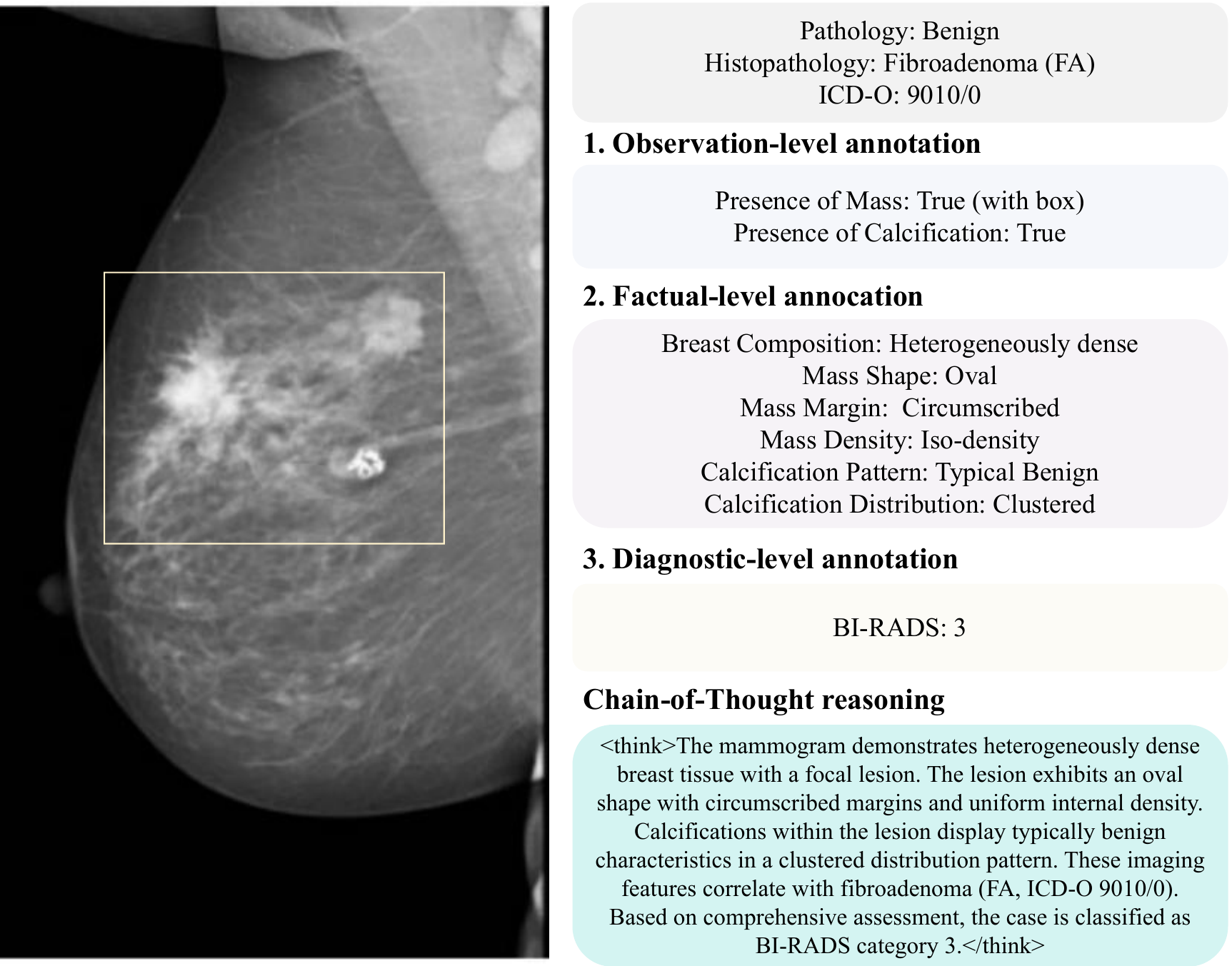} 
    \caption{Case with three levels of annotations in MammoExpert.}
    \label{fig_dataset_example}
\end{figure}

Figure~\ref{fig_dataset_example} illustrates MammoExpert's comprehensive annotation framework through a representative case. The image showcases (1) histopathological subtype and its ICD code, (2) observation of mass and calcification, (3) factual annotations characteristics (mass features, calcification patterns, etc.), (4) diagnostic BI-RADS assessment, and (5) Chain-of-Thought reasoning steps. This labeling structure enables systematic modeling of clinical reasoning pathways.

\subsection{Distribution Statistics}

We split the data using case-level random sampling to keep different cases separate while balancing subtypes. The 1,304 total cases were divided into 1,044 training cases (1,879 images) and 260 test cases (500 images). This method ensures the training and test sets have similar amounts of each cancer type, shown in Figure~\ref{fig_data_statistics}(d). Among benign subtypes, fibroadenoma (FA) and neurofibroma (FN) are the most common, while invasive breast carcinoma (IBC) and ductal carcinoma in situ (DCIS) dominate the malignant category. We carefully balanced subtype distributions during data collection. For rare subtypes, we ensured at least 10 cases per category to maintain diagnostic relevance. A few extremely rare types had fewer samples but were included with all available high-quality images to preserve clinical completeness.

\section{Methodology}
\label{sec4}
\subsection{Task Formulation and Evaluation Metrics}
We formulate breast cancer diagnosis as a binary malignancy classification task. For input mammogram $\mathbf{X} \in \mathbb{R}^{H \times W}$, the model outputs malignancy probability $\hat{y} \in [0,1]$ through feature encoder $\phi(\cdot)$ and decision layer $\psi(\cdot)$:

\begin{equation}
    \hat{y} = \sigma\big(\psi(\phi(\mathbf{X}))\big)
\end{equation}
where $\sigma$ denotes the sigmoid function. Standard classifiers are evaluated using AUROC, AUPRC, and accuracy (ACC) with a 0.5 decision threshold, where confidence intervals (CIs) are calculated via 1,000 bootstrap resamples. For vision-language models (VLMs), predictions are extracted as discrete diagnostic labels (e.g., "malignant" or "benign") through regular expression parsing of textual outputs. We compute ACC against histopathology-confirmed labels, with CIs similarly derived from 1,000 bootstrap iterations. The lack of continuous probability estimates in VLMs precludes AUC-based metric calculations requiring threshold-sensitive confidence scores.

\subsection{Design of Chain-of-Thought}
The MammoExpert dataset allows training AI systems to follow radiologists' three-step reasoning. Using the structured annotations, models first learn to detect lesions and calcifications. Then they analyze features like shape, margins, and calcification patterns. Finally, they predict cancer types and BI-RADS scores by combining these features. This step-by-step training helps AI make decisions similar to how doctors review mammograms.

\begin{figure*}[!t]
    \centering
    \includegraphics[width=\textwidth]{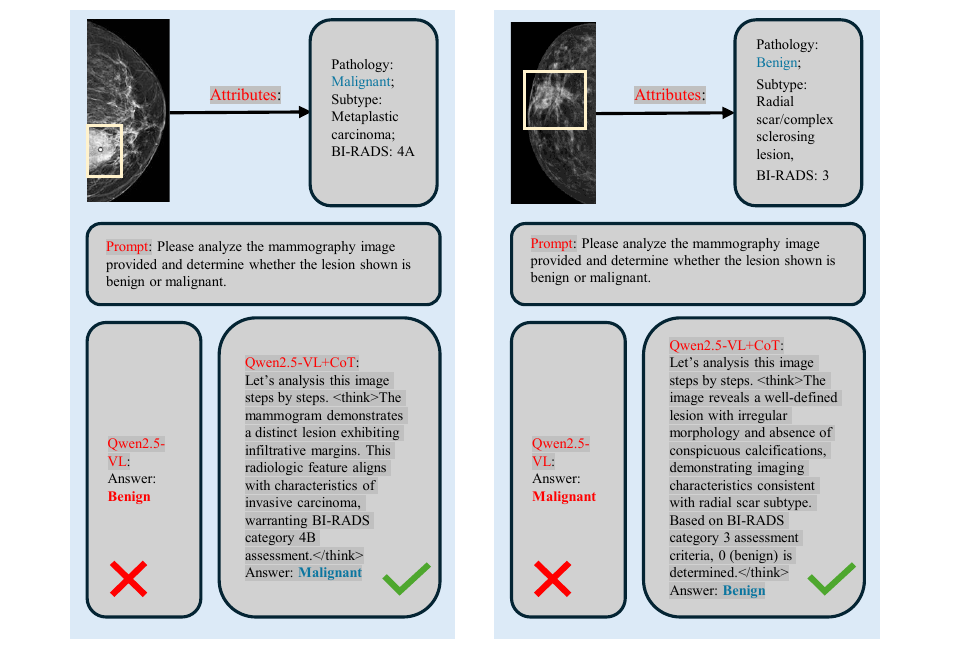}
\caption{Chain-of-Thought Reasoning Validation Cases.}
    \label{fig_cot}
\end{figure*}

For supervised fine-tuning (SFT), we structure the training data to explicitly guide vision-language models (VLMs) through clinically-informed diagnostic workflows. Each training sample consists of a mammogram, immediately followed by a text prompt instructing the model to "Please analyze the mammography image provided and determine whether the lesion shown is benign or malignant." The expected response begins with "Let’s analysis this image steps by steps", and incorporates a <think> section detailing a three-phase analytical procedure: (1) checking for visible lesions/calcifications,
(2) describing their shape/margin features, and (3) combining ob-
servations to assign BI-RADS scores. By explicitly structuring the reasoning chain in this manner, the model is encouraged to follow step-by-step clinical logic rather than producing a single end-point prediction. This structured
format trains models to explicitly follow clinical reasoning patterns. Figure~\ref{fig_cot} visualizes an example of this CoT annotation applied to a training instance, which aligns with the systematic diagnostic process employed by expert radiologists.

\section{Experiments}
\label{sec5}
\subsection{Experimental Setup}
Classification models (ResNet~\cite{he2016deep}, Swin-Transformer~\cite{liu2021Swin}) were trained using $224\!\times\!224$ zero-padded mammograms with five-fold cross validation. Each fold underwent 5-epoch training via AdamW ($\beta_1=0.9$, $\beta_2=0.999$) with batch size 16, initial learning rate $5\!\times\!10^{-5}$ (cosine decay), weight decay 0.1, and milestone-based scheduling at epochs 3--4. The vision-language framework fine-tunes Qwen2.5-VL-3B~\cite{bai2025qwen25vltechnicalreport} on 16$\times$NVIDIA 4090 GPUs using aspect-ratio preserved $224\!\times\!224$ inputs, parameter-efficient tuning with LoRA (rank = 64, $\alpha=64$), constant learning rate $5\!\times\!10^{-5}$, and gradient accumulation (step = 4) over 10,000 training steps.

\subsection{Results on Different Public Datasets}

Incorporating the MammoExpert dataset and the CoT reasoning method substantially improves model performance across multiple mammography datasets. Our approach combining MammoExpert and Qwen2.5-VL reaches 78.04\% accuracy on CBIS-DDSM, which is 7.1\% higher than training only on the original dataset (70.90\%) (Table~\ref{tab:all_datasets_enhanced}). Using MammoExpert data alone already improves CBIS-DDSM accuracy to 74.07\%, while adding reasoning steps provides the full gain. The model trained on INBreast achieves 75.52\% accuracy, which rises to 78.26\% when augmented with MammoExpert data and further improves to 82.44\% with the addition of step-by-step reasoning.
The accuracy grows from 78.62\% to 82.12\% with MammoExpert on the Vindr dataset, and ultimately 85.32\% with the full approach.

These results demonstrate that integrating high-quality medical annotations with structured reasoning significantly enhances model performance and generalization across multiple mammography datasets. The improvements are consistent across all evaluated datasets, highlighting the effectiveness of our framework in leveraging both detailed expert knowledge and reasoning guidance for AI-assisted breast cancer diagnosis.

\begin{table*}[h]
\caption{Performance Comparison on Multiple Datasets.}
\label{tab:all_datasets_enhanced}
\resizebox{\textwidth}{!}{
\begin{tabular}{lccccccccc}
\toprule
\textbf{Model} 
& \multicolumn{3}{c}{\textbf{CBIS-DDSM}} 
& \multicolumn{3}{c}{\textbf{INBreast}} 
& \multicolumn{3}{c}{\textbf{VINDR}} \\ 
\cmidrule(r){2-4} \cmidrule(r){5-7} \cmidrule(r){8-10}
 & AUROC & ACC & Recall & AUROC & ACC & Recall & AUROC & ACC & Recall \\
\midrule
VGGMammo~\cite{shen2019deep} 
& 0.7100 (0.6600, 0.7500) & - & - 
& 0.7352 (0.6902, 0.7753) & - & - 
& 0.7508 (0.7114, 0.7806) & - & - \\

DenseMammo-Single~\cite{quintana2023exploiting} 
& 0.7840 (0.7838, 0.7842) & 0.7030 (0.7016, 0.7044) & 0.7112 (0.7098, 0.7126)
& 0.8052 (0.8026, 0.8110) & 0.7314 (0.7286, 0.7328) & 0.7401 (0.7373, 0.7415)
& 0.8252 (0.8150, 0.8242) & 0.7682 (0.7480, 0.7528) & 0.7735 (0.7533, 0.7781) \\

MorphRes~\cite{wei2022beyond} 
& 0.7958 (0.7958, 0.7958) & - & - 
& 0.8122 (0.8084, 0.8128) & - & - 
& 0.8248 (0.8224, 0.8288) & - & - \\

MorphHR~\cite{wei2022beyond} 
& 0.7964 (0.7964, 0.7964) & - & - 
& 0.8138 (0.8112, 0.8166) & - & - 
& 0.8284 (0.8254, 0.8316) & - & - \\

EfficientMammo~\cite{petrini2022breast} 
& 0.8033 (0.7850, 0.8216) & 0.7554 (0.7554, 0.7554) & 0.7610 (0.7610, 0.7610)
& 0.8226 (0.8084, 0.8426) & 0.7712 (0.7680, 0.7742) & 0.7784 (0.7752, 0.7814)
& 0.8352 (0.8218, 0.8532) & 0.7934 (0.7882, 0.7944) & 0.7991 (0.7939, 0.8001) \\

DenseMammo-Multi~\cite{quintana2023exploiting} 
& 0.8090 (0.8085, 0.8095) & 0.7300 (0.7110, 0.7490) & 0.7366 (0.7176, 0.7556)
& 0.8256 (0.8122, 0.8362) & 0.7506 (0.7421, 0.7636) & 0.7579 (0.7494, 0.7709)
& 0.8408 (0.8312, 0.8506) & 0.7824 (0.7772, 0.7992) & 0.7880 (0.7828, 0.8048) \\

MedCLIP~\cite{wang2022medclip}
& 0.8125 (0.8045, 0.8205) & 0.7010 (0.6560, 0.7460) & 0.7135 (0.6685, 0.7585)
& 0.8285 (0.8205, 0.8365) & 0.7232 (0.7016, 0.7562) & 0.7314 (0.7098, 0.7644)
& 0.8435 (0.8355, 0.8515) & 0.7442 (0.7192, 0.7772) & 0.7501 (0.7251, 0.7831) \\

BioViL~\cite{bannur2023learning}
& 0.8180 (0.8102, 0.8258) & 0.7112 (0.7098, 0.7126) & 0.7194 (0.7180, 0.7208)
& 0.8340 (0.8262, 0.8418) & 0.7396 (0.7368, 0.7410) & 0.7483 (0.7455, 0.7497)
& 0.8490 (0.8412, 0.8568) & 0.7764 (0.7562, 0.7610) & 0.7817 (0.7615, 0.7863) \\

PLIP~\cite{huang2023visual}
& 0.8245 (0.8164, 0.8326) & 0.7234 (0.7234, 0.7234) & 0.7310 (0.7310, 0.7310)
& 0.8415 (0.8334, 0.8496) & 0.7512 (0.7512, 0.7512) & 0.7588 (0.7588, 0.7588)
& 0.8555 (0.8474, 0.8636) & 0.7816 (0.7816, 0.7816) & 0.7873 (0.7873, 0.7873) \\

MedVInT~\cite{moor2023med}
& 0.8310 (0.8228, 0.8392) & 0.7316 (0.7316, 0.7316) & 0.7388 (0.7388, 0.7388)
& 0.8480 (0.8398, 0.8562) & 0.7584 (0.7584, 0.7584) & 0.7659 (0.7659, 0.7659)
& 0.8620 (0.8538, 0.8702) & 0.7888 (0.7888, 0.7888) & 0.7945 (0.7945, 0.7945) \\

Med-Flamingo~\cite{zhang2024development}
& 0.8385 (0.8305, 0.8465) & 0.7436 (0.7436, 0.7436) & 0.7512 (0.7512, 0.7512)
& 0.8565 (0.8485, 0.8645) & 0.7654 (0.7622, 0.7684) & 0.7726 (0.7694, 0.7756)
& 0.8695 (0.8615, 0.8775) & 0.7914 (0.7862, 0.7924) & 0.7971 (0.7919, 0.7981) \\

Qwen2.5-VL-3B~\cite{bai2025qwen25vltechnicalreport} 
& - & 0.7090 (0.6640, 0.7540) & 0.7215 (0.6765, 0.7665)
& - & 0.7552 (0.7336, 0.7882) & 0.7632 (0.7416, 0.7962)
& - & 0.7862 (0.7612, 0.8142) & 0.7914 (0.7664, 0.8194) \\

Qwen2.5-VL-3B\scriptsize{+CoT}
& - & 0.7241 (0.6791, 0.7691) & 0.7368 (0.6918, 0.7818)
& - & 0.7683 (0.7467, 0.8013) & 0.7754 (0.7538, 0.8084)
& - & 0.8035 (0.7785, 0.8315) & 0.8097 (0.7847, 0.8377) \\
\midrule
 & \multicolumn{9}{c}{\textit{Combining MammoExpert}} \\ 
\hline
Qwen2.5-VL-3B~\cite{bai2025qwen25vltechnicalreport} 
& - & \textbf{0.7407} (0.6984, 0.7857) & \textbf{0.7562} (0.7139, 0.8012)
& - & \textbf{0.7826} (0.7403, 0.8288) & \textbf{0.7895} (0.7472, 0.8357)
& - & \textbf{0.8212} (0.7821, 0.8636) & \textbf{0.8267} (0.7876, 0.8691) \\

Qwen2.5-VL-3B\scriptsize{+CoT} 
& - & \textbf{0.7804} (0.7434, 0.8254) & \textbf{0.8013} (0.7643, 0.8463)
& - & \textbf{0.8244} (0.7724, 0.8732) & \textbf{0.8310} (0.7790, 0.8798)
& - & \textbf{0.8532} (0.8142, 0.8868) & \textbf{0.8596} (0.8206, 0.8932) \\
\bottomrule
\end{tabular}
}
\end{table*}

\subsection{Ablation Study on MammoExpert Dataset}
The annotations in MammoExpert enable systematic validation of reasoning-aware frameworks. Table~\ref{tab:main_results} presents two validation cases where the model failed to produce a correct classification using direct inference but successfully arrived at the right conclusion after implementing CoT reasoning. The Qwen2.5-VL-3B model with CoT reasoning demonstrates superior clinical alignment through 4\% accuracy gain over image-only baselines, which is exemplified in Figure~\ref{fig_cot}. This is because the CoT reasoning enables the model to generate sequential textual explanations that condition subsequent predictions, effectively decomposing the classification problem into a series of sub-decisions. This decomposition is particularly advantageous in the MammoExpert setting, which involves classification across 67 histopathology subtypes with significant inter-class visual overlap and substantial class imbalance. The CoT reasoning reduces reliance on global visual shortcuts and mitigates the tendency to over-predict high-frequency classes under uncertainty. And the textual reasoning chain also acts as an auxiliary supervision signal, regularizing the model during training by enforcing alignment between visual cues and domain-specific language patterns. This alignment improves the discriminative performance and enhances interpretability, as the generated rationales provide a transparent window into the model’s decision-making process. In clinical contexts, such interpretability is critical for fostering trust and facilitating expert review, especially when dealing with rare or ambiguous cases. The moderate absolute accuracy (75.7\%) reflects the inherent challenge of diagnosing 67 histopathology subtypes, including rare and ambiguous cases that require expert-level differentiation.

\begin{table*}[!t]
\centering
\caption{Ablation study on MammoExpert Dataset.}
\label{tab:main_results}
\resizebox{\textwidth}{!}{
\begin{tabular}{llccccc} 
\toprule
\textbf{Type} & \textbf{Model} & \textbf{Reasoning} & \textbf{AUROC} & \textbf{AUPRC} & \textbf{ACC} & \textbf{Recall} \\
\midrule
\multirow{4}{*}{CNNs} 
    & ResNet-34  & \xmark & 0.7202 (0.6430, 0.8083) & 0.8073 (0.7308, 0.8878) & 0.6832 (0.6188, 0.7525) & 0.6932 (0.6288, 0.7625) \\
    & ResNet-50  & \xmark & 0.7139 (0.6400, 0.7984) & 0.8108 (0.7360, 0.8858) & 0.6287 (0.5644, 0.6980) & 0.6387 (0.5744, 0.7080) \\
    & ResNet-101 & \xmark & 0.6791 (0.6007, 0.7605) & 0.7860 (0.7023, 0.8657) & 0.6485 (0.5842, 0.7178) & 0.6585 (0.5942, 0.7278) \\
    & ResNet-152 & \xmark & \textbf{0.7394} (0.6670, 0.8135) & \textbf{0.8209} (0.7440, 0.8987) & 0.7129 (0.6535, 0.7723) & 0.7229 (0.6635, 0.7823) \\
\midrule
\multirow{2}{*}{Vision Transformers} 
    & Swin-B & \xmark & 0.7340 (0.6658, 0.8098) & \textbf{0.8547} (0.7991, 0.9216) & 0.6782 (0.6188, 0.7426) & 0.6882 (0.6288, 0.7526) \\
    & Swin-L & \xmark & 0.7272 (0.6499, 0.8036) & 0.8159 (0.7425, 0.8913) & 0.6980 (0.6386, 0.7624) & 0.7080 (0.6486, 0.7724) \\
\midrule
\multirow{3}{*}{Vision--Language Models} 
    & Qwen2.5-VL-3B        & \xmark & - & - & 0.7162 (0.6703, 0.7622) & 0.7262 (0.6803, 0.7722) \\
    & Qwen2.5-VL-3B + CoT  & \cmark & - & - & 0.7568 (0.7108, 0.8000) & 0.7668 (0.7208, 0.8100) \\
    & Qwen2.5-VL-3B + MBCoT & \cmark & - & - & \textbf{0.8088} (0.7612, 0.8486) & \textbf{0.8194} (0.7724, 0.8594) \\
\bottomrule
\end{tabular}%
}
\end{table*}

\begin{table}[!t]
\scriptsize
\centering
\caption{CoT performance on morphologically similar but pathologically different subtypes}
\label{tab:cot_confusable}
\resizebox{0.48\textwidth}{!}{
\begin{tabular}{lccc}
\toprule
\textbf{Subtype}                  & \textbf{Setting}  & \textbf{ACC}  & \textbf{Recall} \\
\midrule
Mucinous carcinoma       & w/o CoT  & 0.8319 & 0.8021 \\
                         & w/ CoT   & \textbf{0.8623} & \textbf{0.8214} \\
\midrule
Invasive ductal carcinoma& w/o CoT  & 0.9018 & 0.8706 \\
                         & w/ CoT   & \textbf{0.9303} & \textbf{0.9021} \\
\bottomrule
\end{tabular}}
\end{table}

To verify whether CoT helps the model discriminate fine-grained differences,
we evaluate several representative easily-confused rare subtype pairs ($i.e.,$ mucinous carcinoma vs invasive ductal carcinoma).
As shown in Table~\ref{tab:cot_confusable}, CoT consistently improves ACC and Recall on these similar-appearing but different subtypes. This confirms that CoT encourages the model to focus on clinically critical fine-grained features (e.g., margin, internal texture, boundary, calcification distribution), which enhances the ability to distinguish pathologically different subtypes with similar imaging features.

\subsection{Rare subtype analysis}

We conduct separate performance analysis on rare subtypes (n $\leq$ 20 cases) and clinically similar but pathologically distinct subtypes (e.g., mucinous carcinoma vs invasive ductal carcinoma), which is shown in Table~\ref{tab:rare_subtypes_all}. The results clearly show that our model achieves stable performance even on extremely rare subtypes, and CoT reasoning brings consistent improvements in distinguishing morphologically similar but pathologically different categories.
The consistent gains of CoT across these difficult cases demonstrate that the model learns generalizable diagnostic features (shape, margin, density, calcification distribution, etc.) guided by clinical reasoning. This is further supported by the radiologist-reviewed reasoning chains, which align with real clinical decision logic. Even for extremely rare subtypes with only 5–10 cases, the model still achieves meaningful and clinically reasonable performance (around 61\%–66\%), which confirms the benchmark captures general diagnostic patterns instead of overfitting to individual instances.

\begin{table}[!t]
  \centering
  \caption{Performance on all rare breast histopathological subtypes (n $\le$ 20)}
  \label{tab:rare_subtypes_all}
  \resizebox{0.48\textwidth}{!}{
  \begin{tabular}{lccc}
    \toprule
    \textbf{Subtype Name} & \textbf{\# Samples} & \textbf{ACC (\%)} & \textbf{Recall (\%)} \\
    \midrule
    Breast liposarcoma                       & 20  & 68.8 & 70.5 \\
    MALT lymphoma                            & 20  & 68.5 & 70.1 \\
    Hemangioblastoma                         & 19  & 68.2 & 69.8 \\
    Secretory carcinoma                      & 18  & 67.9 & 69.5 \\
    Follicular lymphoma                      & 16  & 67.3 & 68.9 \\
    Microinvasive carcinoma                  & 15  & 70.2 & 72.5 \\
    Solid papillary carcinoma                & 14  & 68.5 & 69.9 \\
    Invasive micropapillary carcinoma        & 12  & 66.1 & 71.4 \\
    Atypical lobular hyperplasia             & 10  & 68.1 & 70.4 \\
    Tall cell carcinoma with reversed polarity & 10 & 67.6 & 69.9 \\
    Mucoepidermoid carcinoma                  & 8   & 65.2 & 66.6 \\
    Burkitt lymphoma                          & 8   & 64.8 & 65.1 \\
    Li--Fraumeni syndrome (TP53-related)      & 6   & 63.0 & 64.5 \\
    Papillary adenoma                         & 5   & 61.8 & 62.3 \\
    Peutz--Jeghers syndrome                  & 5   & 60.1 & 61.2 \\
    \bottomrule
  \end{tabular}}
\end{table}

\subsection{Visualization on MammoExpert Dataset}

\begin{figure*}[!t]
\begin{tabular}{
    >{\raggedright\arraybackslash\scriptsize}p{2.5cm} 
    *{6}{c}                                  
}

VGGMammo
    & \parbox[c][3.5cm][c]{2.1cm}{\centering
        \includegraphics[width=2cm, height=2cm]{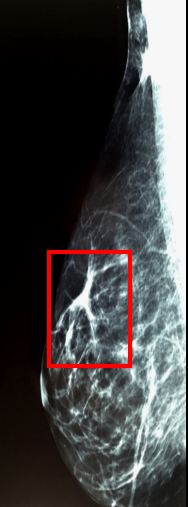}\\
        {\scriptsize pred: Benign \\ gt: Benign}
      }
    & \parbox[c][3.5cm][c]{2.1cm}{\centering
        \includegraphics[width=2cm, height=2cm]{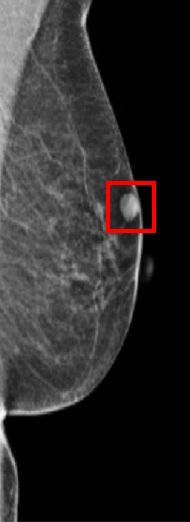}\\
        {\scriptsize pred: Benign \\ gt: Benign}
      }
    & \parbox[c][3.5cm][c]{2.1cm}{\centering
        \includegraphics[width=2cm, height=2cm]{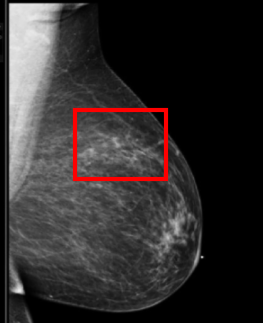}\\
        {\scriptsize \textcolor{red}{pred: Benign \\ gt: Malignant}}
      }
    & \parbox[c][3.5cm][c]{2.1cm}{\centering
        \includegraphics[width=2cm, height=2cm]{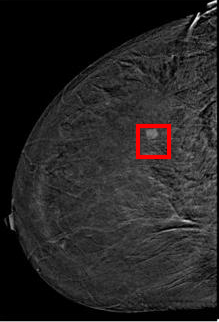}\\
        {\scriptsize pred: Benign \\ gt: Benign}
      }
    & \parbox[c][3.5cm][c]{2.1cm}{\centering
        \includegraphics[width=2cm, height=2cm]{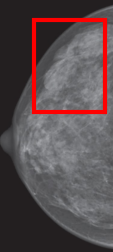}\\
        {\scriptsize \textcolor{red}{pred: Benign \\ gt: Malignant}}
      }
    & \parbox[c][3.5cm][c]{2.1cm}{\centering
        \includegraphics[width=2cm, height=2cm]{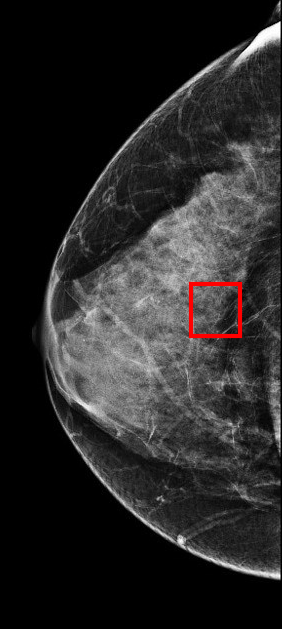}\\
        {\scriptsize pred: Benign \\ gt: Benign}
      }
\\[5mm]

EfficientMammo
    & \parbox[c][3.5cm][c]{2.1cm}{\centering
        \includegraphics[width=2cm, height=2cm]{img/00076-benign.png}\\
        {\scriptsize pred: Benign \\ gt: Benign}
      }
    & \parbox[c][3.5cm][c]{2.1cm}{\centering
        \includegraphics[width=2cm, height=2cm]{img/00096-benign.png}\\
        {\scriptsize pred: Benign \\ gt: Benign}
      }
    & \parbox[c][3.5cm][c]{2.1cm}{\centering
        \includegraphics[width=2cm, height=2cm]{img/00744-malignant.png}\\
        {\scriptsize pred: Malignant \\ gt: Malignant}
      }
    & \parbox[c][3.5cm][c]{2.1cm}{\centering
        \includegraphics[width=2cm, height=2cm]{img/00824-benign.png}\\
        {\scriptsize pred: Benign \\ gt: Benign}
      }
    & \parbox[c][3.5cm][c]{2.1cm}{\centering
        \includegraphics[width=2cm, height=2cm]{img/01167-malignant.png}\\
        {\scriptsize \textcolor{red}{pred: Benign \\ gt: Malignant}}
      }
    & \parbox[c][3.5cm][c]{2.1cm}{\centering
        \includegraphics[width=2cm, height=2cm]{img/02216-benign.png}\\
        {\scriptsize pred: Benign \\ gt: Benign}
      }
\\[5mm]

Qwen2.5-VL-3B
    & \parbox[c][3.5cm][c]{2.1cm}{\centering
        \includegraphics[width=2cm, height=2cm]{img/00076-benign.png}\\
        {\scriptsize pred: Benign \\ gt: Benign}
      }
    & \parbox[c][3.5cm][c]{2.1cm}{\centering
        \includegraphics[width=2cm, height=2cm]{img/00096-benign.png}\\
        {\scriptsize pred: Benign \\ gt: Benign}
      }
    & \parbox[c][3.5cm][c]{2.1cm}{\centering
        \includegraphics[width=2cm, height=2cm]{img/00744-malignant.png}\\
        {\scriptsize \textcolor{red}{pred: Benign \\ gt: Malignant}}
      }
    & \parbox[c][3.5cm][c]{2.1cm}{\centering
        \includegraphics[width=2cm, height=2cm]{img/00824-benign.png}\\
        {\scriptsize pred: Benign \\ gt: Benign}
      }
    & \parbox[c][3.5cm][c]{2.1cm}{\centering
        \includegraphics[width=2cm, height=2cm]{img/01167-malignant.png}\\
        {\scriptsize pred: Malignant \\ gt: Malignant}
      }
    & \parbox[c][3.5cm][c]{2.1cm}{\centering
        \includegraphics[width=2cm, height=2cm]{img/02216-benign.png}\\
        {\scriptsize pred: Benign \\ gt: Benign}
      }
\\[5mm]

Qwen2.5-VL-3B+CoT
    & \parbox[c][3.5cm][c]{2.1cm}{\centering
        \includegraphics[width=2cm, height=2cm]{img/00076-benign.png}\\
        {\scriptsize pred: Benign \\ gt: Benign}
      }
    & \parbox[c][3.5cm][c]{2.1cm}{\centering
        \includegraphics[width=2cm, height=2cm]{img/00096-benign.png}\\
        {\scriptsize pred: Benign \\ gt: Benign}
      }
    & \parbox[c][3.5cm][c]{2.1cm}{\centering
        \includegraphics[width=2cm, height=2cm]{img/00744-malignant.png}\\
        {\scriptsize pred: Malignant \\ gt: Malignant}
      }
    & \parbox[c][3.5cm][c]{2.1cm}{\centering
        \includegraphics[width=2cm, height=2cm]{img/00824-benign.png}\\
        {\scriptsize pred: Benign \\ gt: Benign}
      }
    & \parbox[c][3.5cm][c]{2.1cm}{\centering
        \includegraphics[width=2cm, height=2cm]{img/01167-malignant.png}\\
        {\scriptsize pred: Malignant \\ gt: Malignant}
      }
    & \parbox[c][3.5cm][c]{2.1cm}{\centering
        \includegraphics[width=2cm, height=2cm]{img/02216-benign.png}\\
        {\scriptsize pred: Benign \\ gt: Benign}
      }
\\

\end{tabular}

\caption{Visualization results of four baseline models on MammoExpert test dataset. Each row corresponds to a different model. The four representative models are VGGMammo, EfficientMammo, Qwen2.5-VL-3B, and Qwen2.5-VL-3B+CoT, respectively. For each model, six representative test samples are shown with predicted labels (pred) and ground truth labels (gt). 
Misclassifications are highlighted in red. VGGMammo makes two errors, EfficientMammo and Qwen2.5-VL-3B each make one, while Qwen2.5-VL-3B+CoT correctly classifies all cases. 
These visualizations demonstrate the progressive improvement in diagnostic accuracy of our CoT model.}
\label{fig:four_cases_vertical_leftcaption}

\end{figure*}

Figure~\ref{fig:four_cases_vertical_leftcaption} illustrates visual comparisons across different models evaluated on the MammoExpert dataset. Incorporation of the MammoExpert dataset substantially improves lesion classification performance due to its high-quality annotation and histopathological subtype information, which enables the models to capture subtle imaging patterns that are otherwise ambiguous in less granular datasets. The CoT framework demonstrates superior capacity to resolve critical misclassifications that persist in traditional end-to-end models. In complex cases where morphological indicators (e.g., spiculated margins) conflict with density profiles, direct inference often falters. The CoT model addresses this by constructing a structured reasoning path, which ensures a more robust classification. These explicit reasoning pathways prevent the model from defaulting to majority class shortcuts. They facilitate a more discriminative analysis of nuanced features, which leads to accurate subtype classification even in highly ambiguous cases. Figure~\ref{fig:four_cases_vertical_leftcaption} highlights the superiority of the CoT model in handling the long-tail distribution of subtypes. While baseline models frequently mislabel rare cases due to their visual resemblance to common forms, the CoT model effectively rectifies these errors by validating each diagnostic hypothesis through a structured reasoning pathway. By integrating fine-grained morphological analysis with diagnostic logic, the CoT method provides a more robust reasoning path that successfully separates visually ambiguous classes like mucinous versus invasive ductal carcinoma. It aligns closely with real-world diagnostic workflows of radiologists, which improves the clinical credibility of lesion classification.

\section{Limitations and Ethical Considerations}
\label{sec6}
\subsection{Limitations}

Although MammoExpert includes rich structured annotations, its overall scale is inherently constrained by the limited availability of publicly released mammography data, particularly for rare pathological subtypes. In clinical practice, these subtypes are intrinsically low-frequency and are seldom represented in open-access datasets, which makes large-scale collection challenging and results in class imbalance within the dataset. Such imbalance may restrict the effectiveness of end-to-end training for data-intensive models and reduce their ability to acquire robust feature representations for rare lesion subtypes. The inclusion of fine-grained radiographic attributes and reasoning annotations allows us to partially mitigate this limitation by providing complementary supervisory signals beyond standard image-label pairs. We will further expand the dataset and explore additional strategies to enhance the learning of rare lesion subtypes in the future.

\subsection{Ethical Considerations}

All images in the MammoExpert dataset are obtained from publicly accessible platforms. The use of each sample strictly adheres to the permissions and ethical guidelines specified by the original data providers, which ensures full compliance with ethical standards. Each image is linked to a unique identifier, which prevents any association with personal information. All images are stored on secure servers with access restricted to authorized personnel for data curation. The dataset is designed to support research in a ethical and legally compliant manner, which maintains rigorous standards for patient privacy and data security.

\section{Conclusion}
\label{sec7}
MammoExpert introduces a mammography dataset consisting of 2,379 images annotated with 42 radiographic features and 67 WHO-classified pathological subtypes, which enables AI systems to better emulate radiologists’ diagnostic reasoning. Extensive experiments demonstrate the effectiveness of the constructed dataset. Incorporating MammoExpert into CBIS-DDSM improves diagnostic accuracy by 7.1\% compared to using CBIS-DDSM alone. Furthermore, introducing CoT reasoning yields an additional 4\% performance gain, which highlights the benefits of step-by-step clinical logic in breast cancer detection. Consistent improvements are also observed on the INBreast and VinDr-Mammo datasets, where the full framework achieves accuracy gains of 6.9\% and 6.7\%, respectively. The visualization results show that incorporating the MammoExpert dataset leads to a clear improvement in the classification performance. These findings highlight the value of detailed CoT annotations in MammoExpert, which provide subtype labels and structured diagnostic reasoning. Such annotation design is important for developing diagnostic systems that are reliable in lesion classification and interpretable from a clinical perspective. Future work will extend this framework to other medical imaging modalities while preserving consistency with expert clinical reasoning.

\section{Acknowledgement}
\label{sec8}

This work is supported by National Science Foundation of China (NSFC92470123, NSFC62276005) and the State Key Laboratory of General Artificial Intelligence.

\bibliographystyle{ACM-Reference-Format}
\bibliography{sample-base}

\end{document}